\documentstyle[aclap]{article}

\def\asustackrel#1#2#3{\mathrel{\mathop{#2}\limits_{#3}^{#1}}}
\newcommand\Der[2]{\asustackrel{#1}{\Rightarrow}{#2}}
\newcommand\Lder[2]{\asustackrel{#1}{\Rightarrow}{\ell, #2}}
\newcommand\set[1]{\{ #1 \}}
\newcommand\push[2]{\asustackrel{#1}{\prec}{#2}}
\newcommand\pop[2]{\asustackrel{#1}{\succ}{#2}}
\newcommand\pushpop[1]{\asustackrel{}{\mathrel{\prec \! \succ}}{#1}}
\newcommand\spine[1]{\asustackrel{#1}{\approx}{}}

\newtheorem{definition}{Definition}
\newtheorem{property}{Property}

\title{\vspace{-0.5in}Another Facet of LIG Parsing}
\author{Pierre Boullier \\
INRIA-Rocquencourt \\ 
BP 105 \\
78153 Le Chesnay Cedex, France \\{\tt Pierre.Boullier@inria.fr}}

\begin{document}
\bibliographystyle{fullname}
\maketitle
\vspace{-0.5in}
\begin{abstract}
In this paper\protect\footnotemark\footnotetext{See~\cite{Boullier:RR96} for an
extended version.} we present a new parsing algorithm for linear indexed
grammars (LIGs) in the same spirit as the one described
in~\cite{Vijay-Shanker+Weir:EACL93} for tree adjoining grammars.  For a LIG $L$
and an input string $x$ of length $n$, we build a non ambiguous context-free
grammar whose sentences are all (and exclusively) valid derivation sequences in
$L$ which lead to $x$.  We show that this grammar can be built in ${\cal
O}(n^6)$ time and that individual parses can be extracted in linear time with
the size of the extracted parse tree.  Though this ${\cal O}(n^6)$ upper bound
does not improve over previous results, the average case behaves much better.
Moreover, practical parsing times can be decreased by some statically performed
computations.
\end{abstract}

\section{Introduction}

The class of mildly context-sensitive languages can be described by several
equivalent grammar types.  Among these types we can notably cite tree adjoining
grammars (TAGs) and linear indexed grammars (LIGs).  In
\cite{Vijay-Shanker+Weir:ACL94} TAGs are transformed into equivalent LIGs.
Though context-sensitive linguistic phenomena seem to be more naturally
expressed in TAG formalism, from a computational point of view, many authors
think that LIGs play a central role and therefore the understanding of LIGs and
LIG parsing is of importance.  For example, quoted from
\cite{Schabes+Shieber:ACL94} ``The LIG version of TAG can be used for
recognition and parsing.  Because the LIG formalism is based on augmented
rewriting, the parsing algorithms can be much simpler to understand and easier
to modify, and no loss of generality is incurred''.  In
\cite{Vijay-Shanker+Weir:EACL93} LIGs are used to express the derivations of a
sentence in TAGs.  In \cite{Vijay-Shanker+Weir+Rambow:IWPT95} the approach used
for parsing a new formalism, the D-Tree Grammars (DTG), is to translate a DTG
into a Linear Prioritized Multiset Grammar which is similar to a LIG but uses
multisets in place of stacks.

LIGs can be seen as usual context-free grammars (CFGs) upon which constraints
are imposed.  These constraints are expressed by stacks of symbols associated
with non-terminals.  We study parsing of LIGs, our goal being to define a
structure that verifies the LIG constraints and codes all (and exclusively)
parse trees deriving sentences.

Since derivations in LIGs are constrained CF derivations, we can think of a
scheme where the CF derivations for a given input are expressed by a shared
forest from which individual parse trees which do not satisfied the LIG
constraints are erased.  Unhappily this view is too simplistic, since the
erasing of individual trees whose parts can be shared with other valid trees
can only be performed after some unfolding (unsharing) that can produced a forest
whose size is exponential or even unbounded.

In \cite{Vijay-Shanker+Weir:EACL93}, the context-freeness of adjunction in TAGs
is captured by giving a CFG to represent the set of all possible derivation
sequences.  In this paper we study a new parsing scheme for LIGs based upon similar
principles and which, on the other side, emphasizes as \cite{Lang:CIPT91} and
\cite{Lang:CI94}, the use of grammars (shared forest) to represent parse trees
and is an extension of our previous work~\cite{Boullier:IWPT95}.  

This previous paper describes a recognition algorithm for LIGs, but not a
parser.  For a LIG and an input string, all valid parse trees are actually
coded into the CF shared parse forest used by this recognizer, but, on some
parse trees of this forest, the checking of the LIG constraints can possibly
failed.  At first sight, there are two conceivable ways to extend this
recognizer into a parser:

\begin{enumerate}
\item only ``good'' trees are kept;

\item the LIG constraints are [re-]checked while the extraction of valid trees
      is performed.
\end{enumerate}

As explained above, the first solution can produce an unbounded number of
trees.  The second solution is also uncomfortable since it necessitates the
reevaluation on each tree of the LIG conditions and, doing so, we move away
from the usual idea that individual parse trees can be extracted by a simple
walk through a structure.  

In this paper, we advocate a third way which will use (see
section~\ref{s:RLIG}), the same basic material as the one used
in~\cite{Boullier:IWPT95}.  For a given LIG $L$ and an input string $x$, we
exhibit a non ambiguous CFG whose sentences are all possible valid derivation
sequences in $L$ which lead to $x$.  We show that this CFG can be constructed
in ${\cal O}(n^6)$ time and that individual parses can be extracted in time linear
with the size of the extracted tree.

\section{Derivation Grammar and CF Parse Forest}\label{s:SPF} 

In a CFG $G = (V_N, V_T, P, S)$, the {\sl derives} relation $\Der{}{G}$ is the
set $\set{(\sigma B\sigma' ,\sigma\beta\sigma' )\mid B\rightarrow\beta
\in P\wedge V = V_N\cup V_T\wedge\sigma,\sigma'\in V^*}$.  A {\sl
derivation} is a sequence of strings in $V^*$ s.t. the relation derives holds
between any two consecutive strings.  In a {\sl rightmost} derivation, at each
step, the rightmost non-terminal say $B$ is replaced by the right-hand side
(RHS) of a $B$-production.  Equivalently if $\sigma_0\Der{r_1}{G}\ldots{}
\Der{r_n}{G}\sigma_n$ is a rightmost derivation where the relation symbol is
overlined by the production used at each step, we say that $r_1\ldots{} r_n$
is a rightmost $\sigma_0/\sigma_n$-derivation.

For a CFG $G$, the set of its rightmost $S/x$-derivations, where $x\in {\cal
L}(G)$, can itself be defined by a grammar.

\begin{definition}\label{def:rdg}
Let $G = (V_N, V_T, P, S)$ be a CFG, its {\sl rightmost derivation
grammar} is the CFG $D = (V_N, P, P^D, S)$ where $P^D =\set{A_0\rightarrow
A_1\ldots{} A_q r\mid r = A_0\rightarrow w_0 A_1 w_1\ldots{} w_{q-1} A_q
w_q\in P\wedge w_i\in V^*_T\wedge A_j\in V_N}$
\end{definition}

From the natural bijection between $P$ and $P^D$, we can easily prove that
\begin{center}
 ${\cal L}(D)=\set{r_n\ldots{} r_1\mid r_1\ldots{} r_n\mbox{ is a rightmost }
 S/x\mbox{-derivation in } G}$
\end{center}

This shows that the rightmost derivation language of a CFG is also CF.  We will
show in section~\ref{s:RLIG} that a similar result holds for LIGs.

Following~\cite{Lang:CI94}, CF parsing is the intersection of a CFG and a
finite-state automaton (FSA) which models the input string
$x$\protect\footnotemark\footnotetext{if $x = a_1\ldots{} a_n$, the states can
be the integers $0\ldots{} n$, 0 is the initial state, $n$ the unique
final state, and the transition function $\delta$ is s.t. $i\in\delta (i-1,
a_i)$ and $i\in\delta (i,\varepsilon)$.}.  The result of this intersection
is a CFG $G^{x} = (V^{x}_N, V^{x}_T, P^{x}, [S]^{n}_0)$ called a {\sl shared
parse forest} which is a specialization of the initial CFG $G = (V_N, V_T, P,
S)$ to $x$.  Each production $r^j_i\in P^{x}$, is the production $r_i
\in P$ up to some non-terminal renaming.  The non-terminal symbols in $V^{x}_N$
are triples denoted $[A]^q_p$ where $A\in V_N$, and $p$ and $q$ are states.
When such a non-terminal is productive, $[A]^q_p\Der{+}{G^{x}} w$, we have $q
\in\delta (p, w)$.

If we build the rightmost derivation grammar associated with a shared parse
forest, and we remove all its useless symbols, we get a reduced CFG say $D^x$.
The CF recognition problem for $(G,x)$ is equivalent to the existence of an
$[S]^{n}_0$-production in $D^x$.  Moreover, each rightmost $S/x$-derivation in
$G$ is (the reverse of) a sentence in ${\cal L}(D^x)$.  However, this result is
not very interesting since individual parse trees can be as easily extracted
directly from the parse forest.  This is due to the fact that in the CF case, a
tree that is derived (a parse tree) contains all the information about its
derivation (the sequence of rewritings used) and therefore there is no need to
distinguish between these two notions.  Though this is not always the case with non CF
formalisms, we will see in the next sections that a similar approach, when
applied to LIGs, leads to a shared parse forest which is a LIG while it is
possible to define a derivation grammar which is CF.

\section{Linear Indexed Grammars}\label{s:LIG}

An indexed grammar is a CFG in which stack of symbols are associated with
non-terminals. LIGs are a restricted form of indexed grammars in which the
dependence between stacks is such that at most one stack in the RHS of a
production is related with the stack in its LHS. Other non-terminals are
associated with independant stacks of bounded size.

Following~\cite{Vijay-Shanker+Weir:ACL94}

\begin{definition}
$L = (V_N, V_T, V_I, P_L, S)$ denotes a LIG where $V_N$, $V_T$, $V_I$ and
$P_L$ are respectively finite sets of non-terminals, terminals, stack symbols
and productions, and $S$ is the start symbol.
\end{definition}

In the sequel we will only consider a restricted form of LIGs with productions
of the form
\begin{eqnarray*}
P_L & = &\set{A()\rightarrow w}\cup\set{A(..\alpha)\rightarrow\Gamma_1 B(..\alpha')\Gamma_2}
\end{eqnarray*}
where $A,B\in
 V_N$, $w\in V^*_T\wedge 0
\leq |w|\leq 2$, $\alpha\alpha'\in V^*_I\wedge 0
\leq |\alpha\alpha'|\leq 1$ and $\Gamma_1\Gamma_2\in V_T\cup\set{\varepsilon}\cup\set{C()\mid C
\in V_N}$.

An element like $A(.. \alpha)$ is a {\sl primary constituent} while $C()$ is a
{\sl secondary constituent}.  The stack schema $(.. \alpha)$ of a primary
constituent matches all the stacks whose prefix (bottom) part is left
unspecified and whose suffix (top) part is $\alpha$; the stack of a
secondary constituent is always empty.

Such a form has been chosen both for complexity reasons and to decrease the
number of cases we have to deal with.  However, it is easy to see that this
form of LIG constitutes a normal form.

We use $r()$ to denote a production in $P_L$, where the parentheses remind us
that we are in a LIG!

The {\sl CF-backbone} of a LIG is the underlying CFG in which each production is a LIG
production where the stack part of each constituent has been deleted, leaving only the
non-terminal part.  We will only consider LIGs such there is a bijection
between its production set and the production set of its
CF-backbone\protect\footnotemark\footnotetext{$r_p$ and $r_p()$ with the same
index $p$ designate associated productions.}.

We call {\sl object} the pair denoted $A(\alpha)$ where $A$ is a non-terminal
and $(\alpha)$ a stack of symbols.  Let $V_O =\set{A(\alpha)\mid A\in V_N
\wedge\alpha\in V^*_I}$ be the set of objects.  We define on $(V_O\cup
V_T)^*$ the binary relation {\sl derives} denoted $\Der{}{L}$ (the relation
symbol is sometimes overlined by a production):
\begin{eqnarray*}
\Gamma'_1 A(\alpha''\alpha)\Gamma'_2 &\Der{A(..\alpha)\rightarrow
\Gamma_1 B(..\alpha')\Gamma_2}{L} &\Gamma'_1\Gamma_1 B(\alpha''\alpha')
\Gamma_2\Gamma'_2\\
\Gamma'_1 A()\Gamma'_2 &\Der{A()\rightarrow w}{L} &\Gamma'_1 w\Gamma'_2
\end{eqnarray*}

In the first above element we say that the object $B(\alpha''\alpha')$ is the
{\sl distinguished child} of $A(\alpha''\alpha)$, and if $\Gamma_1\Gamma_2 =
C()$, $C()$ is the {\sl secondary object}. A {\sl derivation} $\Gamma_1,\ldots
,\Gamma_i,\Gamma_{i+1},\ldots{} ,\Gamma_l$ is a sequence of strings where
the relation derives holds between any two consecutive strings

The language defined by a LIG $L$ is the set:
\begin{center}
 ${\cal L}(L)=\set{x\mid S()\Der{+}{L} x\wedge x\in V_T^*}$
\end{center}

As in the CF case we can talk of rightmost derivations when the rightmost
object is derived at each step. Of course, many other derivation strategies may
be thought of.  For our parsing algorithm, we need such a particular derives
relation.  Assume that at one step an object derives both a distinguished child
and a secondary object.  Our particular derivation strategy is such that this
distinguished child will always be derived after the secondary object (and its
descendants), whether this secondary object lays to its left or to its right.
This derives relation is denoted $\Lder{}{L}$ and is called {\sl
linear}\protect\footnotemark\footnotetext{linear reminds us that we are in a
LIG and relies upon a linear (total) order over object occurrences in a
derivation.  See \cite{Boullier:RR96} for a more formal definition.}.

A {\sl spine} is the sequence of objects $A_1(\alpha_1)$ $\ldots{}A_i(\alpha_i)$
$A_{i+1}(\alpha_{i+1})\ldots{} A_p(\alpha_p)$ if, there is a derivation in
which each object $A_{i+1}(\alpha_{i+1})$ is the distinguished child of
$A_i(\alpha_i)$ (and therefore the {\sl distinguished descendant} of
$A_j(\alpha_j),1\leq j\leq i$).

\section{Linear Derivation Grammar}\label{s:RLIG}

For a given LIG $L$, consider a linear $S()/x$-derivation

\begin{center} 
$S()\Lder{r_n()}{L}\ldots{}\Lder{r_i()}{L}\ldots{}\Lder{r_1()}{L} x$
\end{center} 

The sequence of productions $r_1()\ldots{} r_i()\ldots{} r_n()$ (considered in
reverse order) is a string in $P^*_L$.  The purpose of this section is to define
the set of such strings as the language defined by some CFG.

Associated with a LIG $L = (V_N, V_T, V_I, P_L, S)$, we first define a bunch of
binary relations which are borrowed from \cite{Boullier:IWPT95}
\begin{eqnarray*}
\pushpop{1} & = & \set{(A,B) \mid A(..) \rightarrow \Gamma_1 B(..) \Gamma_2 \in P_L} \\
\push{\gamma}{1} & = & \set{(A,B) \mid A(..) \rightarrow \Gamma_1 B(.. \gamma) \Gamma_2 \in P_L} \\
\pop{\gamma}{1} & = & \set{(A,B) \mid A(.. \gamma) \rightarrow \Gamma_1 B(..) \Gamma_2 \in P_L} \\
\pushpop{+} & = & \set{(A_1,A_p) \mid A_1() \Der{+}{L} \Gamma_1
A_p() \Gamma_2 \mbox{ and } A_p()\\
& & \mbox{ is a distinguished descendant of }
A_1()}
\end{eqnarray*}

The {\sl 1-level} relations simply indicate, for each production, which
operation can be apply to the stack associated with the LHS non-terminal to get
the stack associated with its distinguished child;  $\pushpop{1}$ indicates
equality, $\push{\gamma}{1}$ the pushing of $\gamma$, and $\pop{\gamma}{1}$ the
popping of $\gamma$.  

If we look at the evolution of a stack along a spine
$A_1(\alpha_1) \ldots{} A_i(\alpha_i) A_{i+1}(\alpha_{i+1}) \ldots{}
A_p(\alpha_p)$, between any two objects one of the following holds:
$\alpha_i=\alpha_{i+1}$, $\alpha_i \gamma=\alpha_{i+1}$, or
$\alpha_i=\alpha_{i+1} \gamma$.  

The $\pushpop{+}$ relation select pairs of non-terminals $(A_1, A_p)$ s.t.
$\alpha_1 = \alpha_p = \varepsilon$ along non trivial spines.

If the relations $\pop{\gamma}{+}$ and $\spine{}$ are defined as
$\pop{\gamma}{+} = \pop{\gamma}{1} \cup \pushpop{+} \pop{\gamma}{1}$ and
$\spine{} = \bigcup_{\gamma \in V_I} \push{\gamma}{1} \pop{\gamma}{+}$, we can
see that the following identity holds
\begin{property}\label{p:rel}
\begin{eqnarray*}
\pushpop{+} & = & \pushpop{1} \cup \spine{} \cup \pushpop{1} \pushpop{+}\cup \spine{} \pushpop{+}
\end{eqnarray*}
\end{property}

In \cite{Boullier:IWPT95} we can found an
algorithm\protect\footnotemark\footnotetext{Though in the referred paper, these
relations are defined on constituents, the algorithm also applies to
non-terminals.} which computes the $\pushpop{+}$, $\pop{\gamma}{+}$ and
$\spine{}$ relations as the composition of $\pushpop{1}$, $\push{\gamma}{1}$
and $\pop{\gamma}{1}$ in ${\cal O}(|V_N|^3)$ time.

\begin{definition}\label{d:LDG}
For a LIG $L = (V_N, V_T, V_I, P_L, S)$, we call {\sl linear
derivation grammar} (LDG) the CFG $D_L$ (or $D$ when $L$ is understood) $D =
(V_N^D, V_T^D, P^D, S^D)$
where

\begin{itemize}
\item $V_N^D = \set{[A] \mid A \in V_N} \cup \set{[A \rho B] \mid A,B \in V_N
      \wedge \rho \in {\cal R}}$, and ${\cal R}$ is the set of relations
      $\set{\push{\gamma}{1}, \pushpop{1}, \pop{\gamma}{1}, \pushpop{+},
      \spine{}, \pop{\gamma}{+}}$\protect\footnotemark\footnotetext{In fact we
      will only use {\sl valid} non-terminals $[A \rho B]$ for which the
      relation $\rho$ holds between $A$ and $B$.}

\item $V_T^D = P_L$

\item $S^D = [S]$

\item Below, $[\Gamma_1\Gamma_2]$ denotes either the non-terminal symbol $[X]$ when
$\Gamma_1\Gamma_2 = X()$ or the empty string $\varepsilon$ when
$\Gamma_1\Gamma_2 \in V^*_T$. $P^D$ is defined as being
\begin{eqnarray}
\lefteqn{} \nonumber \\ &&\!\!\!\!\!\!\!\!\!\!\!\!\!\!\!\!\set{[A] \rightarrow
r() \mid r() = A() \rightarrow w \in P_L}\\ && \!\!\!\!\!\!\!\!\!\!\!\!\!\!\!\!
\!\!\!\!\cup \set{[A] \rightarrow r() [A \pushpop{+} B] \mid \nonumber \\ &&
\;\; r() = B() \rightarrow w \in P_L} \\ && \!\!\!\!\!\!\!\!\!\!\!\!\!\!\!\!
\!\!\!\!\cup\set{[A \pushpop{+} C] \rightarrow [\Gamma_1\Gamma_2] r() \mid
\nonumber \\ && \;\; r() = A(..) \rightarrow \Gamma_1 C(..) \Gamma_2 \in P_L}
\\ && \!\!\!\!\!\!\!\!\!\!\!\!\!\!\!\! \!\!\!\!\cup\set{[A \pushpop{+} C]
\rightarrow [A \spine{} C]} \\ && \!\!\!\!\!\!\!\!\!\!\!\!\!\!\!\!
\!\!\!\!\cup\set{[A \pushpop{+} C] \rightarrow [B \pushpop{+} C]
[\Gamma_1\Gamma_2] r() \mid \nonumber \\ && \;\; r() = A(..) \rightarrow
\Gamma_1 B(..) \Gamma_2 \in P_L} \\ && \!\!\!\!\!\!\!\!\!\!\!\!\!\!\!\!
\!\!\!\!\cup\set{[A \pushpop{+} C] \rightarrow [B \pushpop{+} C] [A \spine{}
B]} \\ && \!\!\!\!\!\!\!\!\!\!\!\!\!\!\!\! \!\!\!\!\cup \set{[A \spine{} C]
\rightarrow [B \pop{\gamma}{+} C] [\Gamma_1\Gamma_2] r() \mid \nonumber \\ &&
\;\; r() = A(..) \rightarrow \Gamma_1 B(..\gamma) \Gamma_2 \in P_L} \\ &&
\!\!\!\!\!\!\!\!\!\!\!\!\!\!\!\! \!\!\!\!\cup \set{[A \pop{\gamma}{+} C]
\rightarrow [\Gamma_1\Gamma_2] r() \mid \nonumber \\ && \;\; r() = A(..\gamma)
\rightarrow \Gamma_1 C(..) \Gamma_2 \in P_L} \\ &&
\!\!\!\!\!\!\!\!\!\!\!\!\!\!\!\! \!\!\!\!\cup\set{[A \pop{\gamma}{+} C]
\rightarrow [\Gamma_1\Gamma_2] r() [A \pushpop{+} B] \mid \nonumber \\ && \;\;
r() = B(..\gamma) \rightarrow \Gamma_1 C(..) \Gamma_2 \in P_L}
\end{eqnarray}
\end{itemize}

\end{definition}

The productions in $P^D$ define all the ways linear derivations can be composed
from linear sub-derivations.  This compositions rely on one side upon
property~\ref{p:rel} (recall that the productions in $P_L$, must be produced in
reverse order) and, on the other side, upon the order in which secondary spines
(the $\Gamma_1 \Gamma_2$-spines) are processed to get the linear derivation
order.

In \cite{Boullier:RR96}, we prove that LDGs are not ambiguous (in fact they are
SLR(1)) and define
\begin{eqnarray*}
  {\cal L} (D) & =  &\set{r_1()\ldots{} r_n()\mid S()\Lder{r_n()}{L}
 \ldots{}\Lder{r_1()}{L} x \\
&& \;\; \wedge x\in {\cal L}(L)}
\end{eqnarray*}

If, by some classical algorithm, we remove from $D$ all its useless
symbols, we get a reduced CFG say $D' = (V_N^{D'}, V_T^{D'}, P^{D'},
S^{D'})$.  In this grammar, all its terminal symbols, which are
productions in $L$, are useful.  
By the way, the construction of
$D'$ solve the emptiness problem for LIGs: $L$ specify the empty set
iff the set $V_T^{D'}$ is empty\protect\footnotemark\footnotetext{In
\cite{Vijay-Shanker+Weir:EACL93} the emptiness problem for LIGs is
solved by constructing an FSA.}.

\section{LIG parsing}\label{s:LIGP}

Given a LIG $L = (V_N, V_T, V_I, P_L, S)$ we want to find all the syntactic
structures associated with an input string $x \in V^*_T$.  In
section~\ref{s:SPF} we used a CFG (the shared parse forest) for representing
all parses in a CFG.  In this section we will see how to build a CFG which
represents all parses in a LIG.

In \cite{Boullier:IWPT95} we give a recognizer for LIGs with the following scheme:
in a first phase a general CF parsing algorithm, working on the CF-backbone
builds a shared parse forest for a given input string $x$.  In a second phase,
the LIG conditions are checked on this forest.  This checking can result in
some subtree (production) deletions, namely the ones for which there is no valid symbol
stack evaluation.  If the resulting grammar is not empty, then x is a sentence.
However, in the general case, this resulting grammar is not a shared parse
forest for the initial LIG in the sense that the computation of stack of
symbols along spines are not guaranteed to be consistent.  Such invalid spines
are not deleted during the check of the LIG conditions because they could be
composed of sub-spines which are themselves parts of other valid spines.  One
way to solve this problem is to unfold the shared parse forest and to extract
individual parse trees.  A parse tree is then kept iff the LIG conditions
are valid on that tree.  But such a method is not practical since the number of
parse trees can be unbounded when the CF-backbone is cyclic.  Even for non
cyclic grammars, the number of parse trees can be exponential in the size of
the input.  Moreover, it is problematic that a worst case polynomial size
structure could be reached by some sharing compatible both with the syntactic
and the ``semantic'' features.

However, we know that derivations in TAGs are context-free (see
\cite{Vijay-Shanker:PhD87}) and \cite{Vijay-Shanker+Weir:EACL93} exhibits a CFG
which represents all possible derivation sequences in a TAG.  We will show that
the analogous holds for LIGs and leads to an ${\cal O}(n^6)$ time parsing
algorithm.

\begin{definition}

Let $L = (V_N, V_T, V_I, P_L, S)$ be a LIG, $G = (V_N, V_T, P_G, S)$ its
CF-backbone, $x$ a string in ${\cal L}(G)$, and $G^x = (V_N^{x}, V_T^{x},
P_G^x, S^{x})$ its shared parse forest for $x$.  We define the {\sl LIGed
forest} for $x$ as being the LIG $L^x = (V_N^{x}, V_T^{x}, V_I, P_L^x, S^{x})$
s.t. $G^x$ is its CF-backbone and its productions are the productions of
$P_G^x$ in which the corresponding stack-schemas of $L$ have been added.
For example $r^q_p() = [A]^k_i(.. \alpha) \rightarrow [B]^j_i(.. \alpha')
[C]^k_j() \in P_L^x$ iff $r^q_p = [A]^k_i \rightarrow [B]^j_i [C]^k_j \in P_G^x
\wedge r_p = A \rightarrow B C \in G \wedge r_p() = A(.. \alpha) \rightarrow
B(.. \alpha') C() \in L$.

\end{definition}

Between a LIG $L$ and its LIGed forest $L^x$ for $x$, we have:
\begin{eqnarray*}
  x \in {\cal L}(L) & \Longleftrightarrow & x \in {\cal L}(L^x)
\end{eqnarray*}

If we follow\cite{Lang:CI94}, the previous definition which produces a LIGed
forest from any $L$ and $x$ is a (LIG)
parser\protect\footnotemark\footnotetext{Of course, instead of $x$, we can
consider any FSA.}: given a LIG $L$ and a string $x$, we have constructed a new
LIG $L^{x}$ for the intersection ${\cal L}(L) \cap \set{x}$, which is the
shared forest for all parses of the sentences in the intersection.  However, we
wish to go one step further since the parsing (or even recognition) problem for
LIGs cannot be trivially extracted from the LIGed forests.

Our vision for the parsing of a string $x$ with a LIG $L$ can be summarized in
few lines.  Let $G$ be the CF-backbone of $L$, we first build $G^x$ the CFG
shared parse forest by any classical general CF parsing algorithm and then
$L^x$ its LIGed forest.  Afterwards, we build the reduced LDG $D_{L^x}$
associated with $L^x$ as shown in section~\ref{s:RLIG}.

The recognition problem for $(L, x)$ (i.e.\  is $x$ an element of ${\cal L}(L)$) is
equivalent to the non-emptiness of the production set of $D_{L^x}$.

Moreover, each linear $S()/x$-derivation in $L$ is (the reverse of) a string in
${\cal L}(D_{L^x})$\protect\footnotemark\footnotetext{In fact, the terminal
symbols in $D_{L^x}$ are productions in $L^x$ (say $R^q_p()$), which trivially
can be mapped to productions in $L$ (here $r_p()$).}.  So the extraction of
individual parses in a LIG is merely reduced to the derivation of strings in a
CFG.

An important issue is about the complexity, in time and space, of $D_{L^x}$.
Let $n$ be the length of the input string $x$.  Since $G$ is in binary form we
know that the shared parse forest $G^x$ can be build in ${\cal O}(n^3)$ time
and the number of its productions is also in ${\cal O}(n^3)$.  Moreover, the
cardinality of $V^x_N$ is ${\cal O}(n^2)$ and, for any given non-terminal, say
$[A]^q_p$, there are at most ${\cal O}(n)$ $[A]^q_p$-productions.  Of course,
these complexities extend to the LIGed forest $L^x$.

We now look at the LDG complexity when the input LIG is a
LIGed forest.  In fact, we mainly have to check two forms of productions (see
definition~\ref{d:LDG}).  The first form is production (6) ($[A \pushpop{+} C]
\rightarrow [B \pushpop{+} C] [A \spine{} B]$), where three different
non-terminals in $V_N$ are implied (i.e.\ $A$, $B$ and $C$), so the number of
productions of that form is cubic in the number of non-terminals and therefore
is ${\cal O}(n^6)$.

In the second form (productions (5), (7) and (9)), exemplified by $[A \spine{}
C] \rightarrow [B \pop{\gamma}{+} C] [\Gamma_1\Gamma_2] r()$, there are four
non-terminals in $V_N$ (i.e.\ $A$, $B$, $C$, and $X$ if $\Gamma_1 \Gamma_2 =
X()$) and a production $r()$ (the number of relation symbols $\pop{\gamma}{+}$
is a constant), therefore, the number of such productions seems to be of fourth degree in
the number of non-terminals and linear in the number of productions.  However,
these variables are not independant.  For a given $A$, the number of triples
$(B, X, r())$ is the number of $A$-productions hence ${\cal O}(n)$.  So, at
the end, the number of productions of that form is ${\cal O}(n^5)$.  

We can easily check that the other form of productions have a lesser degree.

Therefore, the number of productions is dominated by the first form and the
size (and in fact the construction time) of this grammar is ${\cal O}(n^6)$.

This (once again) shows that the recognition and parsing problem for a LIG can
be solved in ${\cal O}(n^6)$ time.

For a LDG $D = (V_N^D, V_T^D, P^D, S^D)$, we note that for any given
non-terminal $A \in V^D_N$ and string $\sigma \in {\cal L}(A)$ with $|\sigma|
\geq 2$, a single production $A \rightarrow X_1 X_2$ or $A \rightarrow X_1 X_2
X_3$ in $P^D$ is needed to ``cut'' $\sigma$ into two or three non-empty pieces
$\sigma_1$, $\sigma_2$, and $\sigma_3$, such that $X_i \Der{*}{D} \sigma_i$,
except when the production form number (4) is used.  In such a case, this
cutting needs two productions (namely (4) and (7)).  This shows that the
cutting out of any string of length $l$, into elementary pieces of length 1, is
performed in using ${\cal O}(l)$ productions.  Therefore, the extraction of a
linear $S()/x$-derivation in $L$ is performed in time linear with the length of
that derivation.  If we assume that the CF-backbone $G$ is non cyclic, the
extraction of a parse is linear in $n$.  Moreover, during an extraction, since
$D_{L^x}$ is not ambiguous, at some place, the choice of another $A$-production
will result in a different linear derivation.

Of course, practical generations of LDGs must improve over a blind application
of definition~\ref{d:LDG}.  One way is to consider a top-down strategy: the
$X$-productions in a LDG are generated iff $X$ is the start symbol or occurs in
the RHS of an already generated production.  The examples in section~\ref{s:e}
are produced this way.

If the number of ambiguities in the initial LIG is bounded, the size of
$D_{L^x}$, for a given input string $x$ of length $n$, is linear in $n$.

The size and the time needed to compute $D_{L^x}$ are closely related to the
actual sizes of the $\pushpop{+}$, $\pop{\gamma}{+}$ and $\spine{}$ relations.
As pointed out in \cite{Boullier:IWPT95}, their ${\cal O}(n^4)$ maximum sizes
seem to be seldom reached in practice.  This means that the average parsing
time is much better than this ${\cal O}(n^6)$ worst case.

Moreover, our parsing schema allow to avoid some useless computations.  Assume
that the symbol $[A\pushpop{+} B]$ is useless in the LDG $D_L$ associated with
the initial LIG $L$, we know that any non-terminal s.t. $[[A]^j_i\pushpop{+}
[B]^l_k]$ is also useless in $D_{L^x}$.  Therefore, the static computation of a
reduced LDG for the initial LIG $L$ (and the corresponding $\pushpop{+}$,
$\pop{\gamma}{+}$ and $\spine{}$ relations) can be used to direct the parsing
process and decrease the parsing time (see section~\ref{s:e}).

\section{Two Examples}\label{s:e} 
\subsection{First Example}
In this section, we illustrate our algorithm with a LIG $L= (\set{S,T}, \set{a,
b, c}, \set{\gamma_a, \gamma_b, \gamma_c}, P_L, S)$ where $P_L$ contains
the following productions:
\[\begin{array}{ll}
r_1() = S(..) \rightarrow S(.. \gamma_a) a &
r_2() = S(..) \rightarrow S(.. \gamma_b) b \\
r_3() = S(..) \rightarrow S(.. \gamma_c) c &
r_4() = S(..) \rightarrow T(..) \\
r_5() = T(.. \gamma_a) \rightarrow a T(..) &
r_6() = T(.. \gamma_b) \rightarrow b T(..) \\
r_7() = T(.. \gamma_c) \rightarrow c T(..) &
r_8() = T() \rightarrow c
\end{array}\]

It is easy to see that its CF-backbone $G$, whose production set $P_G$ is:
\[\begin{array}{llll}
S \rightarrow S a &
S \rightarrow S b & 
S \rightarrow S c &
S \rightarrow T \\
T \rightarrow a T &
T \rightarrow b T &
T \rightarrow c T &
T \rightarrow c
\end{array}\]

\noindent defines the language ${\cal L}(G) = \set{wcw' \mid w, w' \in {\set{a, b,
c}}^*}$.  We remark that the stacks of symbols in $L$ constrain the string $w'$
to be equal to $w$ and therefore the language ${\cal L}(L)$ is $\set{wcw \mid w
\in {\set{a, b, c}}^*}$.

We note that in $L$ the key part is played by the middle $c$, introduced
by production $r_8()$, and that this grammar is non
ambiguous, while in $G$ the symbol $c$, introduced by the last production $T
\rightarrow c$, is only a separator between $w$ and $w'$ and that this grammar
is ambiguous (any occurrence of $c$ may be this separator).

The computation of the relations gives:
\[\begin{array}{lllllll}
&&&&\pushpop{1} & = & \set{(S,T)} \\
\push{\gamma_a}{1} & = & \push{\gamma_b}{1} & = & \push{\gamma_c}{1} & = & \set{(S,S)} \\
\pop{\gamma_a}{1} & = & \pop{\gamma_b}{1} & = & \pop{\gamma_c}{1} & = & \set{(T,T)} \\
&&&&\pushpop{+} & = & \set{(S,T)} \\
&&&&\spine{} & = & \set{(S,T)} \\
\pop{\gamma_a}{+} & = & \pop{\gamma_b}{+} & = & \pop{\gamma_c}{+} & = &
\set{(T,T), (S,T)}
\end{array}\]

The production set $P^D$ of the LDG $D$ associated with $L$ is:
\[\begin{array}{lll@{\hspace{.6in}}r}
[S] & \rightarrow & r_8() [S \pushpop{+} T] & (2) \\{}
[S \pushpop{+} T] & \rightarrow & r_4() & (3) \\{}
[S \pushpop{+} T] & \rightarrow & [S \spine{} T] & (4) \\{}
[S \spine{} T] & \rightarrow & [S \pop{\gamma_a}{+} T] r_1() & (7)
\\{}
[S \spine{} T] & \rightarrow & [S \pop{\gamma_b}{+} T] r_2() & (7)
\\{}
[S \spine{} T] & \rightarrow & [S \pop{\gamma_c}{+} T] r_3() & (7) \\{}
[S \pop{\gamma_a}{+} T] & \rightarrow & r_5() [S \pushpop{+} T] & (9)
\\{}
[S \pop{\gamma_b}{+} T] & \rightarrow & r_6() [S \pushpop{+} T] & (9)
\\{}
[S \pop{\gamma_c}{+} T] & \rightarrow & r_7() [S \pushpop{+} T] & (9)
\end{array}\]

The numbers $(i)$ refer to definition~\ref{d:LDG}.  We can easily
checked that this grammar is reduced.

Let $x=ccc$ be an input string.  Since $x$ is an element of ${\cal L}(G)$, its
shared parse forest $G^x$ is not empty.  Its production set $P_G^x$ is:
\[\begin{array}{ll}
r_3^1 = [S]_{0}^{3}\rightarrow [S]_{0}^{2} c &
r_4^2 = [S]_{0}^{3}\rightarrow [T]_{0}^{3} \\{}
r_3^3 = [S]_{0}^{2}\rightarrow [S]_{0}^{1} c &
r_4^4 = [S]_{0}^{2}\rightarrow [T]_{0}^{2} \\{}
r_4^5 = [S]_{0}^{1}\rightarrow [T]_{0}^{1} &
r_7^6 = [T]_{0}^{3}\rightarrow c[T]_{1}^{3} \\{}
r_7^7 = [T]_{1}^{3}\rightarrow c[T]_{2}^{3} &
r_8^8 = [T]_{2}^{3}\rightarrow c \\{}
r_7^9 = [T]_{0}^{2}\rightarrow c[T]_{1}^{2} &
r_8^{10} = [T]_{1}^{2}\rightarrow c \\{}
r_8^{11} = [T]_{0}^{1}\rightarrow c &
\end{array}\]

\noindent We can observe that this shared parse forest denotes in fact three
different parse trees. Each one corresponding to a different cutting out of $x=
wcw'$ (i.e.\ $w=\varepsilon $ and $w'=cc$, or $w=c$ and $w'=c$, or $w=cc$ and
$w'=\varepsilon$). 

The corresponding LIGed forest whose start symbol is $S^x=[S]_{0}^{3}$ and
production set $P_L^x$ is:
\[\begin{array}{lllll}
r_3^1() & = & [S]_{0}^{3}(..) & \rightarrow &  [S]_{0}^{2}(.. \gamma_c) c \\{}
r_4^2() & = & [S]_{0}^{3}(..) & \rightarrow &  [T]_{0}^{3}(..) \\{}
r_3^3() & = & [S]_{0}^{2}(..) & \rightarrow &  [S]_{0}^{1}(.. \gamma_c) c \\{}
r_4^4() & = & [S]_{0}^{2}(..) & \rightarrow &  [T]_{0}^{2}(..) \\{}
r_4^5() & = & [S]_{0}^{1}(..) & \rightarrow &  [T]_{0}^{1}(..) \\{}
r_7^6() & = & [T]_{0}^{3}(.. \gamma_c) & \rightarrow &  c[T]_{1}^{3}(..) \\{}
r_7^7() & = & [T]_{1}^{3}(.. \gamma_c) & \rightarrow &  c[T]_{2}^{3}(..) \\{}
r_8^8() & = & [T]_{2}^{3}() & \rightarrow &  c \\{}
r_7^9() & = & [T]_{0}^{2}(.. \gamma_c) & \rightarrow &  c[T]_{1}^{2}(..) \\{}
r_8^{10}() & = & [T]_{1}^{2}() & \rightarrow &  c \\{}
r_8^{11}() & = & [T]_{0}^{1}() & \rightarrow &  c
\end{array}\]

For this LIGed forest the relations are:
\begin{eqnarray*}
\pushpop{1} & = & \set{([S]_{0}^{3},[T]_{0}^{3}), ([S]_{0}^{2},[T]_{0}^{2}), ([S]_{0}^{1},[T]_{0}^{1})} \\
\push{\gamma_c}{1} & = & \set{([S]_{0}^{3},[S]_{0}^{2}), ([S]_{0}^{2},[S]_{0}^{1})} \\
\pop{\gamma_c}{1} & = & \set{([T]_{0}^{3},[T]_{1}^{3}), ([T]_{1}^{3},[T]_{2}^{3}), ([T]_{0}^{2},[T]_{1}^{2})} \\
\spine{} & = & \set{([S]_{0}^{3},[T]_{1}^{2})} \\
\pushpop{+} & = & \pushpop{1} \cup \spine{} \\
\pop{\gamma_c}{+} & = & \pop{\gamma_c}{1} \cup \set{([S]_{0}^{3},[T]_{1}^{3}), ([S]_{0}^{2},[T]_{1}^{2})}
\end{eqnarray*}

The start symbol of the LDG associated with the LIGed
forest $L^x$ is $[[S]^3_0]$.  If we assume that an $A$-production is generated
iff it is an $[[S]^3_0]$-production or $A$ occurs in an already generated
production, we get:
\[\begin{array}{lll@{\hspace{.3in}}r}
[[S]^3_0] & \rightarrow & r_8^{10}() [[S]^3_0 \pushpop{+} [T]_{1}^{2}] & (2) \\{}
[[S]^3_0 \pushpop{+} [T]_{1}^{2}] & \rightarrow & [[S]^3_0 \spine{} [T]_{1}^{2}] & (4) \\{}
[[S]^3_0 \spine{} [T]_{1}^{2}] & \rightarrow & [[S]^2_0 \pop{\gamma_c}{+}
[T]_{1}^{2}] r_3^1() & (7) \\{}
[[S]^2_0 \pop{\gamma_c}{+} [T]_{1}^{2}] & \rightarrow & r_7^9() [[S]^2_0
\pushpop{+} [T]_{0}^{2}] & (9)\\{}
[[S]^2_0 \pushpop{+} [T]_{0}^{2}] & \rightarrow & r_4^4()  & (3)
\end{array}\]

This CFG is reduced.  Since its production set is non empty, we have $ccc \in {\cal L}(L)$.
Its language is $\set{r_8^{10}() r_7^9() r_4^4() r_3^1()}$ which shows that the
only linear derivation in $L$ is $S() \Lder{r_3()}{L} S(\gamma_c) c
\Lder{r_4()}{L} T(\gamma_c) c \Lder{r_7()}{L} c T() c \Lder{r_8()}{L} ccc$.

In computing the relations for the initial LIG $L$, we remark that though $T
\pop{\gamma_a}{+} T$, $T \pop{\gamma_b}{+} T$, and $T \pop{\gamma_c}{+} T$, the
non-terminals $[T
\pop{\gamma_a}{+} T]$, $[T \pop{\gamma_b}{+} T]$, and $[T \pop{\gamma_c}{+} T]$
are not used in $P^D$.  This means that for any LIGed forest $L^x$, the
elements of the form $([T]^q_p, [T]^{q'}_{p'})$ do not need to be computed in
the $\pop{\gamma_a}{+} $, $\pop{\gamma_b}{+}$ , and $\pop{\gamma_c}{+} $
relations since they will never produce a useful non-terminal.  In this
example, the subset $\pop{\gamma_c}{1}$ of $\pop{\gamma_c}{+}$ is useless.

The next example shows the handling of a cyclic grammar.

\subsection{Second Example}
The following LIG $L$, where $A$ is the start symbol:
\[\begin{array}{ll}
r_1() = A(..) \rightarrow A(.. \gamma_a) &
r_2() = A(..) \rightarrow B(..) \\
r_3() = B(.. \gamma_a) \rightarrow B(..) &
r_4() = B() \rightarrow a
\end{array}\]

\noindent is cyclic (we have $A \Der {+}{} A$ and $B \Der {+}{} B$ in its
CF-backbone), and the stack schemas in production $r_1()$ indicate that an
unbounded number of push $\gamma_a$ actions can take place, while production
$r_3()$ indicates an unbounded number of pops.  Its CF-backbone is unbounded
ambiguous though its language contains the single string $a$.

The computation of the relations gives:
\begin{eqnarray*}
\pushpop{1} & = & \set{(A,B)} \\
\push{\gamma_a}{1} & = & \set{(A,A)} \\
\pop{\gamma_a}{1} & = & \set{(B,B)} \\
\pushpop{+} & = & \set{(A,B)} \\
\spine{} & = & \set{(A,B)} \\
\pop{\gamma_a}{+} & = & \set{(A,B), (B,B)}
\end{eqnarray*}

The start symbol of the LDG associated with $L$ is $[A]$ and its productions
set $P^D$ is:
\[\begin{array}{lll@{\hspace{.6in}}r}
[A] & \rightarrow & r_4() [A \pushpop{+} B] & (2) \\{}
[A \pushpop{+} B] & \rightarrow & r_2() & (3) \\{}
[A \pushpop{+} B] & \rightarrow & [A \spine{} B] & (4) \\{}
[A \spine{} B] & \rightarrow & [A \pop{\gamma_a}{+} B] r_1() & (7)\\{}
[A \pop{\gamma_a}{+} B] & \rightarrow & r_3() [A \pushpop{+} B] & (9)
\end{array}\]

We can easily checked that this grammar is reduced.

We want to parse the input string $x=a$ (i.e.\ find all the linear
$S()/a$-derivations).

Its LIGed forest, whose start symbol is $[A]_{0}^{1}$ is:
\[\begin{array}{lllll}
r_1^1() & = & [A]_{0}^{1}(..)  & \rightarrow & [A]_{0}^{1}(.. \gamma_a) \\
r_2^2() & = & [A]_{0}^{1}(..)  & \rightarrow & [B]_{0}^{1}(..) \\
r_3^3() & = & [B]_{0}^{1}(.. \gamma_a)  & \rightarrow & [B]_{0}^{1}(..) \\
r_4^4() & = & [B]_{0}^{1}() & \rightarrow & a
\end{array}\]

For this LIGed forest $L^x$, the relations are:
\begin{eqnarray*}
\pushpop{1} & = & \set{([A]_{0}^{1},[B]_{0}^{1})} \\
\push{\gamma_a}{1} & = & \set{([A]_{0}^{1},[A]_{0}^{1})} \\
\pop{\gamma_a}{1} & = & \set{([B]_{0}^{1},[B]_{0}^{1})} \\
\spine{} & = & \set{([A]_{0}^{1},[B]_{0}^{1})} \\
\pushpop{+} & = & \set{([A]_{0}^{1},[B]_{0}^{1})} \\
\pop{\gamma_a}{+} & = & \set{([A]_{0}^{1},[B]_{0}^{1}), ([B]_{0}^{1},[B]_{0}^{1})}
\end{eqnarray*}

The start symbol of the LDG associated with $L^x$ is $[[A]_{0}^{1}]$.  If we
assume that an $A$-production is generated iff it is an
$[[A]_{0}^{1}]$-production or $A$ occurs in an already generated production,
its production set is: \[\begin{array}{lll@{\hspace{.3in}}r} [[A]_{0}^{1}] &
\rightarrow &r_4^4() [[A]_{0}^{1} \pushpop{+} [B]_{0}^{1}] & (2) \\{}
[[A]_{0}^{1} \pushpop{+} [B]_{0}^{1}] & \rightarrow & r_2^2() & (3) \\{}
[[A]_{0}^{1} \pushpop{+} [B]_{0}^{1}] & \rightarrow & [[A]_{0}^{1} \spine{}
[B]_{0}^{1}] & (4) \\{} [[A]_{0}^{1} \spine{} [B]_{0}^{1}] & \rightarrow &
[[A]_{0}^{1} \pop{\gamma_a}{+} [B]_{0}^{1}] r_1^1() & (7)\\{} [[A]_{0}^{1}
\pop{\gamma_a}{+} [B]_{0}^{1}] & \rightarrow & r_3^3() [[A]_{0}^{1} \pushpop{+}
[B]_{0}^{1}] & (9)
\end{array}\]

This CFG is reduced.  Since its production set is non empty, we have $a \in
{\cal L}(L)$.  Its language is $\set{r_4^4() \set{r_3^3()}^k r_2^2() \set{r_1^1()}^k \mid
0 \leq k}$ which shows that the only valid linear derivations w.r.t. $L$ must
contain an identical number $k$ of productions which push $\gamma_a$ (i.e.\
the production $r_1()$) and productions which pop $\gamma_a$ (i.e.\ the
production $r_3()$).

As in the previous example, we can see that the element $[B]_{0}^{1}
\pop{\gamma_a}{+} [B]_{0}^{1}$ is useless.

\section{Conclusion} We have shown that the parses of a LIG can be represented
by a non ambiguous CFG.  This representation captures the fact that the values
of a stack of symbols is well parenthesized.  When a symbol $\gamma$ is pushed
on a stack at a given index at some place, this very symbol must be popped some
place else, and we know that such (recursive) pairing is the essence of
context-freeness.

In this approach, the number of productions and the construction time of this
CFG is at worst ${\cal O}(n^6)$, though much better results occur in practical
situations.  Moreover, static computations on the initial LIG may decrease this
practical complexity in avoiding useless computations.  Each sentence in this
CFG is a derivation of the given input string by the LIG, and is extracted in
linear time.

\end{document}